\documentclass[10pt,twocolumn,letterpaper]{article}

\usepackage{iccv}
\usepackage{times}
\usepackage{epsfig}
\usepackage{graphicx}
\usepackage{amsmath}
\usepackage{amssymb}
\usepackage{booktabs}
\usepackage{multirow}
\usepackage{nicefrac}
\usepackage[table]{xcolor}
\usepackage{makecell}
\newlength\savewidth
\newcommand{\tablestyle}[2]{\setlength{\tabcolsep}{#1}\renewcommand{\arraystretch}{#2}\centering\footnotesize}
\renewcommand{\paragraph}[1]{\vspace{1.2mm}\noindent\textbf{#1}}
\newcolumntype{x}[1]{>{\centering\arraybackslash}p{#1pt}}
\newcolumntype{y}[1]{>{\raggedright\arraybackslash}p{#1pt}}
\newcolumntype{z}[1]{>{\raggedleft\arraybackslash}p{#1pt}}
\usepackage[breaklinks=true,bookmarks=false,colorlinks,citecolor=citecolor]{hyperref}
\definecolor{citecolor}{HTML}{0071bc}

\iccvfinalcopy 

\def\iccvPaperID{****} 

\ificcvfinal\fi

\begin{document}

\title{SimFLE: Simple Facial Landmark Encoding \\ for Self-Supervised Facial Expression Recognition in the Wild}

\author{Jiyong Moon\\
Dongguk University\\
{\tt\small asdwldyd@dgu.ac.kr}
\and
Seongsik Park\thanks{Corresponding author}\\
Dongguk University\\
{\tt\small s.park@dgu.ac.kr}
}

\maketitle
\ificcvfinal\thispagestyle{empty}\fi

\begin{abstract}
One of the key issues in facial expression recognition in the wild (FER-W) is that curating large-scale labeled facial images is challenging due to the inherent complexity and ambiguity of facial images. Therefore, in this paper, we propose a self-supervised simple facial landmark encoding (SimFLE) method that can learn effective encoding of facial landmarks, which are important features for improving the performance of FER-W, without expensive labels. Specifically, we introduce novel FaceMAE module for this purpose. FaceMAE reconstructs masked facial images with elaborately designed semantic masking. Unlike previous random masking, semantic masking is conducted based on channel information processed in the backbone, so rich semantics of channels can be explored. Additionally, the semantic masking process is fully trainable, enabling FaceMAE to guide the backbone to learn spatial details and contextual properties of fine-grained facial landmarks. Experimental results on several FER-W benchmarks prove that the proposed SimFLE is superior in facial landmark localization and noticeably improved performance compared to the supervised baseline and other self-supervised methods. The code is available at: \url{https://github.com/jymoon0613/simfle}.
\end{abstract}

\section{Introduction}
\label{sec:intro}
\begin{figure}[t]
  \centering
   \includegraphics[width=1\linewidth]{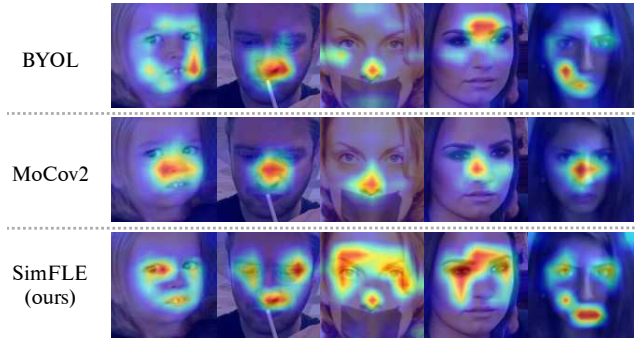}

   \caption{Attention visualization with Grad-CAM~\cite{ETCV2}. Compared to other self-supervised methods, the proposed SimFLE is better at localizing facial landmarks and capturing their associations.}
   \label{fig1}
\end{figure}
Facial expression recognition in the wild (FER-W) is a task in which computers perform emotion recognition by understanding facial expressions in natural and uncontrolled environments~\cite{FETCS1, FETCS2}. FER-W is essential for building emotion-aware intelligent systems (\eg social robots, smart mirrors) because facial expression is one of the most powerful signals of human being~\cite{ETCE1, FETCE1}.

For better performance in FER-W, it is important to effectively encode the decisive fine-grained facial features, or \textit{facial landmarks}\footnote{In this paper, the term facial landmark is used in the sense of a decisive
facial part that represents facial expressions.}, such as eyes and mouth of a given facial image~\cite{FETCL1, FETCL2}. Most recent studies have used visual attention mechanism to exploit facial landmarks~\cite{FS3, FS7, FS9, FS12}. The proposed attention module filters out unnecessary information and highlights fine-grained facial features in a given facial image. Additionally, an image crop method based on the position of facial landmarks has been proposed~\cite{FS2, FS5, FS14}. The crop method explicitly extracts the position of the facial landmarks through a detection algorithm, and exploits the facial landmarks by cropping the image in patch units based on the detected position.

Although these approaches promise improved performance, they have one major limitation: most of the methods has focused on supervised learning, which requires a large amount of labeled data for training. The visual complexity of real-world facial images requires a significant amount of time and high-level expertise to annotate~\cite{FNS2, FETCD3, FETCD4}. Additionally, the inherent ambiguity of facial expressions and the annotation bias coming from different levels of expertise require rigorous verification processes for label reliability.~\cite{FETCD2}. This makes curating of large-scale labeled facial images challenging and prompts training methods that are less label-dependent.

To this end, we propose a simple facial landmark encoding (SimFLE) method that learns effective encoding of fine-grained facial landmarks in an \textit{unsupervised} way. Recently, self-supervised learning has shown promising results in obtaining semantic representations even without labeled data~\cite{SSLP1, SSLC1, SSLC3, SSLS2}. Inspired by these results, we aimed to design a self-supervised method that is most specialized for FER-W. 

Our SimFLE is divided into two branches. First, the backbone learns global facial features common to a given facial domain through contrastive method~\cite{SSLC2, SSLC6, SSLC7} in a global facial feature learning (GFL) branch. At the same time, the backbone learns details of fine-grained facial landmarks through a fine-grained facial feature learning (FFL) branch. Specifically, we propose novel FaceMAE module, a masked autoencoder~\cite{ETCM2} specialized for FER-W, to achieve the goal of FFL. FaceMAE is trained to reconstruct masked facial images with elaborately designed semantic masking. Semantic masking is different from previous random masking in that masking is performed based on channel information encoded by the backbone and that it is trainable. These simple settings allows the backbone to learn details of fine-grained facial landmarks at the backbone level without heavy attention modules, crop-based landmark extraction, or even class labels. FaceMAE also excels at learning overall facial context and associations between facial landmarks. It should be noted that the modules of SimFLE (\ie FaceMAE) are only used during pre-training stage, so the single backbone is used for downstream tasks. 

Our main contributions can be summarized as follows:

\vspace{1mm}\noindent (1) We propose SimFLE, an end-to-end self-supervised learning method for FER-W. With SimFLE, the backbone network can learn representations of decisive facial landmarks and their associations without expensive labels.

\vspace{1mm}\noindent (2) We designed a novel FaceMAE module. FaceMAE's channel-based trainable semantic masking and autoencoding process is key to learning the spatial and contextual properties of facial landmarks. Fig.~\ref{fig1} shows the superiority of our method in learning facial landmarks.

\vspace{1mm}\noindent (3) We evaluate the proposed SimFLE on four FER-W benchmarks: AffectNet, RAF-DB, FERPlus, and SFEW. In all evaluation setups, SimFLE outperforms other self-supervised methods and also significantly closes the gap with supervised state-of-the-arts.
\section{Related Work} 
\label{sec:related}
\begin{figure*}[t]
  \centering
   \includegraphics[width=1\linewidth]{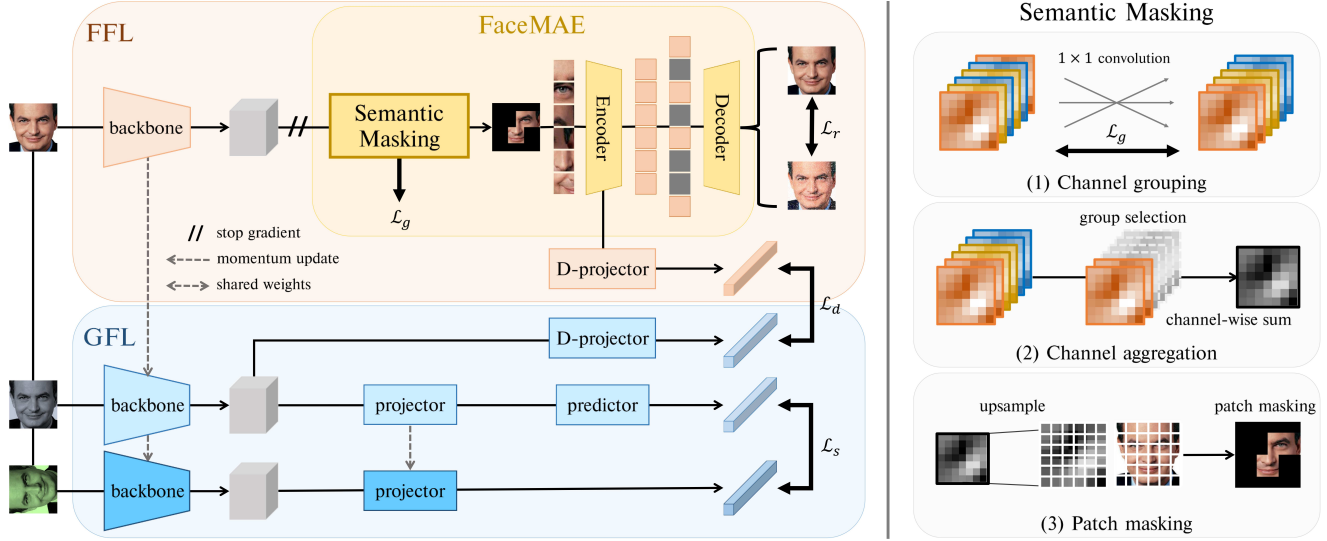}

   \caption{Overall architecture of the proposed SimFLE (left) and pipeline of semantic masking (right). SimFLE is divided into two branches: GFL and FFL. In GFL, the backbone CNN learns global facial features through contrastive method. In FFL, the backbone CNN learns fine-grained facial landmarks with FaceMAE module. First, the feature maps output by the backbone from the input facial image are entered into FaceMAE. The output feature maps are grouped at the channel level according to the similarity of the spatial region they are focused on. One of the channel groups is selected, and channels belonging to the selected channel group are aggregated into one feature map through a channel-wise sum. The aggregated feature map is divided into patches of fixed size, and the patches are masked in ascending order of average pixel values according to a defined masking ratio. FaceMAE's autoencoder reconstructs the semantically masked facial image. Finally, GFL and FFL are connected indirectly through knowledge distillation.}
   \label{fig2}
\end{figure*}
\subsection{Facial Expression Recognition in the Wild}
\label{subsec:fer}
Many methods have been proposed for FER-W~\cite{FS3, FS4, FS6, FS8, FS11, FS13}. For better performance, it is essential to exploit fine-grained facial landmarks. Therefore, attention-based methods~\cite{FS3, FS7, FS9, FS12} or crop-based methods~\cite{FS2, FS5, FS14} have been mainly proposed. The visual attention mechanism removes useless features and emphasizes decisive features~\cite{ETCA1, ETCA2}. In FER-W, an attention network was proposed to focus on fine-grained facial regions while ignoring unnecessary background~\cite{FS3}. Similarly, a region-based attention network was proposed to capture local facial features to solve the occlusion and variant poses of FER-W~\cite{FS5}. These attention networks extended to multi-head attention structures~\cite{FS7}. Additionally, Transformer architecture based on the self-attention mechanism has been actively applied for salient feature selection~\cite{FS12, FS18, FS19, FS20}. On the other hand, the crop-based method finds facial landmarks in the pre-processing stage, and the image patch cropped based on the facial landmarks is used as an input for the feature extractor~\cite{FS14}. Pre-trained landmark detectors were used to locate facial landmarks~\cite{FS2}. These crop-based methods were also used as part of attention networks~\cite{FS5}.

Similar to the above methods, our SimFLE focuses on fine-grained facial landmarks of a given facial image. However, we do not use landmark-based crop methods or complicated attention mechanisms. Additionally, we achieve the goal in an unsupervised way that without any labels, which is different from previous semi-supervised approaches~\cite{FNS1, FNS2}.
\subsection{Self-Supervised Learning}
\label{subsec:ssl}
Self-supervised learning provides effective representations for downstream tasks without requiring labels~\cite{SSLS2}. Many self-supervised methods based on pretext tasks~\cite{SSLP1, SSLP2, SSLP3}, generation~\cite{SSLG1, SSLG2}, clustering~\cite{SSLR1, SSLR2} have been proposed. Among them, contrastive methods~\cite{SSLC2, SSLC5, SSLC6, SSLC7} show state-of-the-art performance especially in image recognition. The contrastive method learns invariant global features by encouraging augmented views of the same input to have more similar representations compared to augmented views of different inputs~\cite{SSLS1, SSLC4}. Besides, masked autoencoding is one of the generative self-supervised methods that trains to reconstruct randomly masked images. Masked autoencoding has strengths in context learning because a defined autoencoder infers the entire image with only limited information~\cite{SSLG2, ETCM2}.

Similar to the multi-task approach~\cite{SSLM1, SSLM2, SSLM3}, our SimFLE jointly uses two different types of self-supervised methods. SimFLE learns global facial features using a contrastive method. At the same time, SimFLE learns fine-grained facial landmarks with masked autoencoding, which is one of the generative methods.
\section{Our Method}
\label{sec:method}
The overall architecture of SimFLE is shown in Fig.~\ref{fig2}. The architecture is divided into two branches for different purposes: global facial feature learning (GFL) and fine-grained facial feature learning (FFL). The two branches are connected via distillation. We adopt convolutional neural networks (CNN) as our backbone structure. More details are described as follows.
\subsection{Global Facial Feature Learning}
\label{subsec:gfl}
Before considering fine-grained features, it is important for the backbone to learn common features that exist globally in a given domain. In GFL, the contrastive method is used to learn the invariant global representations at the image-level. Inspired by BYOL~\cite{SSLC3}, GFL consists of two CNN backbones, two projectors, and one predictor. The entire structure can be divided into an online stream $f_{\theta}$ and a target stream $f_{\rho}$, respectively, and only the online stream has a predictor.

Let $I\in{\mathbb{R}^{h\times{w}\times{c}}}$ denote the input facial image where $h$, $w$, and $c$ refer to the height, width, and the number of channels, respectively. Given $I$, sets of random augmentations $t_1$ and $t_2$ is applied to generate augmented views $v_1$ and $v_2$. The two views fed into two streams are encoded into representations $f_{\theta}\left(v_1\right)$ and $f_{\rho}\left(v_2\right)$, respectively. Finally, the similarity loss is calculated from the two representations:
\begin{equation}
\mathcal{L}_{s}=2-2\cdot\left(\frac{f_{\theta}\left(v_1\right)\cdot{f_{\rho}\left(v_2\right)}}{\lVert{f_{\theta}\left(v_1\right)}\rVert_2\cdot\lVert{f_{\rho}\left(v_2\right)}\rVert_2}\right).
\label{equation1}
\end{equation}
\noindent This loss term is the mean squared error of the $\ell_2$ normalized online representation and target representation. The same process can be done again by swapping the positions of $v_1$ and $v_2$. Through the similarity loss, the backbone learns global facial features that should exist in a given facial domain while learning similarity between two augmented views from the same image. It should be noted that the $f_{\rho}$ is a exponential moving average~\cite{SSLC3, SSLC5} of the $f_{\theta}$: the update on loss is applied only to the online stream parameters $\theta$, and the target stream parameters $\rho$ is updated through $\rho\leftarrow\tau\rho+\left(1-\tau\right)\theta$, where $\tau\in[0,1]$.
\subsection{Fine-Grained Facial Feature Learning}
\label{subsec:ffl}
As mentioned earlier, fine-grained facial features, or \textit{facial landmarks}, such as eyes and mouth are essential to improving the performance of FER-W~\cite{FETCL1, FETCL2}. In FFL, we allow the backbone to learn effective encoding of these facial landmarks without the help of expensive labels. Specifically, we propose novel FaceMAE module to achieve the goal of FFL.

\paragraph{FaceMAE.} The masked autoencoding is useful for context learning because the entire  image is inferred from limited visible information~\cite{SSLG2, ETCE3}. However, we designed FaceMAE, a masked autoencoder~\cite{ETCM2} specialized for FER-W, by making two major modifications to the original masked autoencoding scheme. With FaceMAE, the backbone can automatically learn the spatial details of facial landmarks and their associations beyond learning the overall facial context. The two modifications are that (1) FaceMAE uses elaborately designed semantic masking to mask facial images based on semantic channel information, and (2) the semantic masking process is trainable, so the FaceMAE can refine the masking results at every iteration.

\paragraph{Channel-based Semantic Masking.} Similar to the literature of fine-grained visual classification (FGVC)~\cite{ETCF1, ETCF2, ETCF3, ETCF4, ETCF5}, we mainly considers the channel information of feature maps to detect fine-grained facial landmarks. Specifically, we newly designed a masking strategy based on channel information to explore the rich semantics of channels.

First, the feature maps output by the backbone from the input facial image $I$ are entered into FaceMAE. Let $X\in{\mathbb{R}^{{h^\prime}\times{w^\prime}\times{c^\prime}}}$ denote the output feature map where $h^\prime$, $w^\prime$, and $c^\prime$ refer to the height, width, and the number of channels, respectively. Each channel in the feature map will have its unique spatial features, but will have similarities in terms of the local region they are focused on. Therefore, channel grouping~\cite{ETCM3, ETCM5} is conducted to remove redundant information and consider globality. Specifically, The $c^\prime$ channels constituting the $X$ is grouped into pre-defined $N_g$ channel groups according to their spatial similarity. Channel grouping can be implemented through a $1\times1$ convolution layer, and grouping results are optimized through channel grouping loss:
\begin{equation}
\mathcal{L}_{g}=\sum\limits_{\substack{{i,j}\\{g_{i}}\neq{g_{j}}}}{\text{D}\left(C_{i},C_{j}\right)}^2-\sum\limits_{\substack{{i,j}\\{g_{i}}={g_{j}}}}{\text{D}\left(C_{i},C_{j}\right)}^2,
\label{equation2}
\end{equation}

\noindent where $0\leq{i,j}\leq{c^\prime}$, and $g_i$ and $g_j$ refer to the channel groups to which channels $C_{i}$ and $C_{j}$ belong, respectively. $\text{D}\left(C_{i},C_{j}\right)$ is the cosine similarity between channels $C_i$ and $C_j$ defined as $\frac{C_i\cdot{C_j}}{\lVert{C_i}\rVert_2\cdot\lVert{C_j}\rVert_2}$.

After channel grouping, one of the channel groups is randomly selected, and $\nicefrac{c^\prime}{N_g}$ channels of the selected group $\widetilde{g}$ are aggregated into score map $X_{\widetilde{g}}\in{\mathbb{R}^{{h^\prime}\times{w^\prime}}}$ through a channel-wise sum. Then, the score map $X_{\widetilde{g}}$ is upsampled to be equal to the height and width of the input image, and it is divided into $N_p$ patches $\mathcal{P}=[P_1, P_2,\dots,P_{N_{p}}]$, where $P_{i}\in{\mathbb{R}^{{s}\times{s}}}$. For masking, the average pixel value of each patch is calculated:
\begin{equation}
a_{i}=\frac{1}{s^2}\sum_{x,y}P_{i}(x,y),
\label{equation3}
\end{equation}
\noindent where $0\leq{x,y}\leq{s}$. 

Finally, the patches are masked in ascending order of $\mathcal{A}=[a_1, a_2,\dots, a_{N_{p}}]$, according to the pre-defined masking ratio $\gamma$. The resulting masked image $m$ is reconstructed by the autoencoder. The reconstruction loss $\mathcal{L}_{r}$ is the pixel-level mean squared error between the reconstructed image $\hat{I}$ and original image $I$. We compute the loss only for the $N_m$ masked patches:
\begin{equation}
\mathcal{L}_{r}=\frac{1}{N_m}\sum_{i}^{N_m}\left({I(P_{i})-\hat{I}(\hat{P}_{i})}\right)^2.
\label{equation4}
\end{equation}

\begin{figure}[t]
  \centering
   \includegraphics[width=1\linewidth]{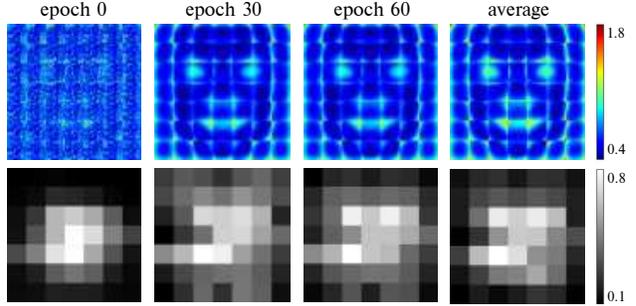}

   \caption{Average pixel-level reconstruction loss of random masking (top) and average patch-level masking position of semantic masking (bottom) over 60 epochs. FaceMAE struggles with reconstruction of facial landmarks, so semantic masking is gradually adjusted to not mask those regions.}
   \label{fig3}
\end{figure}
\paragraph{Trainable Refinement Process.} It should be noted that the semantic masking process is trainable according to the reconstruction results. Since FaceMAE has to minimize the reconstruction loss, it will try to avoid masking fine-grained facial landmarks, which are difficult to reconstruct. Fig.~\ref{fig3} shows the validity of this assumption. Fig.~\ref{fig3} compares the average pixel-level reconstruction loss when using random masking (top) and the average patch-level mask position when using semantic masking (bottom) by tracking for 60 epochs. Initial reconstruction loss was similar in all pixel positions, but it gradually became prominent in positions that could be considered facial landmarks, such as the eyes and mouth. This clearly demonstrates that FaceMAE struggles with reconstruction of facial landmarks compared to redundant backgrounds. At the same time, semantic masking is gradually adjusted from the initial state so that these fine-grained regions are not masked (close to 1) and only unnecessary regions are filtered out (close to 0). 

\begin{figure}[t]
  \centering
   \includegraphics[width=1\linewidth]{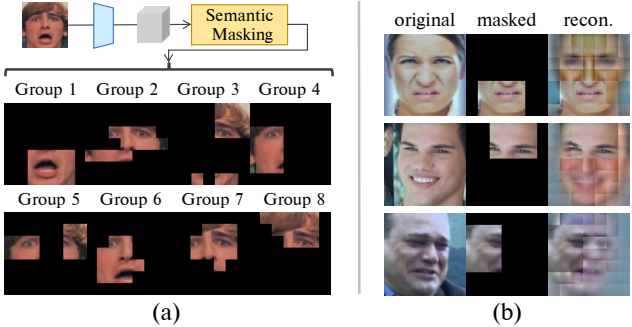}
    
   \caption{Examples of effects when using FaceMAE. (a) Result images masked with 8 randomly selected channel groups from one image. FaceMAE refines features at the backbone level to detect facial landmarks. (b) Examples of masked autoencoding results of FaceMAE. From semantically masked images, FaceMAE conveys effective encodings of facial landmarks and their associations to the backbone.}
   \label{fig4}
\end{figure}

Considering that semantic masking is based on patch-divided channel information, regions of facial landmarks should have higher values than other regions in the output feature map in order not to be masked. This means that the backbone that outputs target feature maps should be trained to clearly detect facial landmarks and encode their representations into each channel. As a result, the backbone adjusts their feature extractors and refines channel semantics to fit fine-grained facial landmarks. As shown in Fig.~\ref{fig4} (a), when masking was conducted based on randomly selected 8 channel groups, all results showed that facial landmarks were preserved as much as possible. This means that the channels of all feature maps were adjusted at the backbone level to focus on fine-grained facial landmarks. Additionally, since the channel group for masking is randomly selected, the backbone can explore distinctive facial landmarks encoded in multiple channel groups at every iteration. 

\paragraph{Facial Context Learning.} Although the iterative refinement process allows the facial landmarks to be maintained at the channel level, a high masking ratio (\ie 0.75) results in some facial landmarks being masked. However, it is useful in that the backbone can learn associations between facial landmarks and the overall context of facial expressions. This is because FaceMAE's encoder needs to encode associations between facial landmarks to predict the masked facial landmarks through some visible ones, as shown in Fig.~\ref{fig4} (b). Hence, the backbone can learn associations of facial landmarks or the overall facial context more deeply by FaceMAE.
\subsection{Connecting Two Branches}
\label{subsec:connecting}
As shown in Fig.~\ref{fig2}, GFL and FFL share the same backbone and are trained jointly. However, training the backbone end-to-end on both branches can make the training unstable because the learning objectives of GFL and FFL are different. Additionally, FaceMAE can easily access the backbone and control the feature extraction through trainable semantic masking process. This means that FaceMAE can cause corruption in the process of adjusting the backbone according to its own interests. Therefore, we avoid direct connection of FFL to the backbone, and connect GFL and FFL indirectly through knowledge distillation~\cite{KDCL1, KDCL2}. GFL and FFL share the learned knowledge through distillation loss defined as follows:
\begin{equation}
\mathcal{L}_{d}=\sum_{i\in\{m,v_1,v_2\}}\text{KL}\left(\psi({z_{i}}/t),~\psi(\bar{z}/t)\right),
\label{equation5}
\end{equation}
\noindent where $z_{i}$ is the encoded representations by the distillation projectors for $i$. $\text{KL}\left(\cdot\right)$ is the Kullback-Leibler divergence loss~\cite{ETCL1}. $\psi\left(\cdot/t\right)$ is the softmax function with temperature $t=4$. $\bar{z}$ is the mean of the three representations.

\paragraph{Total Loss.} The proposed SimFLE method is optimized end-to-end manner. The total loss to be minimized can be derived as:
\begin{equation}
\mathcal{L}_{total}=\mathcal{L}_{s}+\mathcal{L}_{r}+\alpha\mathcal{L}_{d}+\beta\mathcal{L}_{g},
\label{equation6}
\end{equation}
\noindent where $\alpha$ and $\beta$ are the weight parameters for $\mathcal{L}_{d}$ and $\mathcal{L}_{g}$, respectively.  
\section{Experiments}
\label{sec:experiments}
\begin{table*}[t]
\tablestyle{4.5pt}{1.1}
\begin{tabular}{y{50}|x{35}|x{22}x{22}|x{27}x{27}x{27}x{27}|x{22}x{22}|x{30}x{30}x{30}}
\hline
\multicolumn{1}{x{56}|}{\multirow{3}{*}{method}} & \multirow{3}{*}{pre-trained} & \multicolumn{2}{c|}{linear evaluation} & \multicolumn{4}{c|}{semi-supervised learning}                      & \multicolumn{2}{c|}{fine-tuing} & \multicolumn{3}{c}{transfer learning} \\ \cline{3-13} 
\multicolumn{1}{c|}{}                        &                              & AN-8        & AN-7       & AN-8$_{1}$ & AN-8$_{10}$ & AN-7$_{1}$ & AN-7$_{10}$ & AN-8    & AN-7 & RAF-DB      & FER+       & SFEW       \\ \hline \hline
Supervised-IN                                & ImageNet                     & 28.01              & 35.09             & 13.35          & 40.51          & 24.06           & 46.00          & 46.81          & 52.11          & 82.24       & 81.26      & 33.41      \\
Supervised-AN                                & AffectNet                    & -                  & -                 & -                   & -              & -               & -              & -              & -              & 85.23       & 86.07      & 49.42      \\ \hline
SimSiam~\cite{SSLC1}                         & AffectNet                    & 29.68              & 34.80             & 29.61          & 42.91          & 36.06           & 48.97          & 50.79          & 55.97          & 85.53       & 86.58      & 51.51      \\
BYOL~\cite{SSLC3}                            & AffectNet                    & 41.64              & 47.20             & 33.78          & 43.94          & 38.11           & 50.71          & 50.31          & 58.09          & 87.65       & 87.63      & 53.36      \\
MoCo~\cite{SSLC5}                            & AffectNet                    & 37.73              & 44.51             & 30.91          & 43.19          & 38.14           & 50.00          & 47.16          & 53.69          & 80.74       & 83.58      & 41.30      \\
MoCov2~\cite{SSLC6}                          & AffectNet                    & 38.06              & 45.46             & 32.98          & 43.36          & 38.37           & 50.11          & 50.39          & 57.00          & 85.46       & 86.74      & 40.37      \\
SimCLR~\cite{SSLC4}                          & AffectNet                    & 38.41              & 44.97             & 33.76          & 43.04          & 38.80           & 48.74          & 45.11          & 51.80          & 83.15       & 81.48      & 49.19      \\ \hline
SimFLE (ours)                                & AffectNet                    & \textbf{44.01}     & \textbf{50.40}    & \textbf{35.89} & \textbf{46.19} & \textbf{40.23}           & \textbf{53.09}          & \textbf{52.61}          & \textbf{59.14}          & \textbf{88.53}       & \textbf{88.11}      & \textbf{56.38}      \\ \hline
\end{tabular}
\caption{Results of experimental evaluation. Linear evaluation, semi-supervised learning, fine-tuning, and transfer learning results are presented with accuracy (\%). All methods were re-implemented and trained from scratch.}
\label{table1}
\end{table*}
\subsection{Experiment Setup}
\label{subsec:setup}
\paragraph{Datasets.} We experimented on four FER-W benchmark datasets: AffectNet, RAF-DB, FERPlus, and SFEW. \textit{AffectNet}~\cite{FETCD1} is the largest FER-W dataset, consisting of 287,651 training images and 4,000 validation images with 8 emotion classes. Since test images are not provided, we used validation images for test. \textit{RAF-DB}~\cite{FETCD2} is one of the famous FER-W datasets, consisting of 12,271 training images and 3,068 test images. In our experiment, only 7 basic emotion classes are used. \textit{FERPlus}~\cite{FETCD4} is an extended dataset that has been relabeled to overcome the limitations of the existing FER2013~\cite{FETCD5}, containing 25,045 training, 3,191 validation, and 3,137 test images. \textit{SFEW}~\cite{FETCD6} is a static facial expression dataset selected from movies. It contains 891 training, 431 validation, and 372 test images with 7 emotion classes. Since test labels are not provided, we used validation images for test. The detailed data distributions are presented in the Supplement~\ref{app:data}.

\paragraph{Network Architectures.} We used the newly initialized ResNet-50~\cite{ETCM4} as the backbone CNN in our experiments. The two distillation projectors, two GFL projectors, and GFL predictor consist of a fully connected layer with a batch normalization and a ReLU activation, followed by another fully connected layer. In FaceMAE, we set the number of channel groups $N_{g}=32$, the masking ratio $\gamma=0.75$, and the patch size $s=16$. The previously proposed MAE~\cite{ETCM2} with ViT-base~\cite{ETCM1} was used for the autoencoder architecture.

\paragraph{Implementation Details.} For SimFLE training, images were resized to $224\times224$ except for FFL which uses $112\times112$ resolution. For data augmentation, the augmentation list of SimCLR~\cite{SSLC4} was used. We pre-trained the backbone for 100 epochs on AffectNet training set through the proposed SimFLE method, and an SGD optimizer with momentum of 0.9 and weight decay of 0.0001 was used. The batch size was set to 256, and the initial learning rate was set to 0.05. The learning rate had a cosine decay schedule~\cite{ETCE2, SSLC4}. Our proposed method was implemented with the PyTorch framework on three NVIDIA RTX3090 GPUs.
\subsection{Experimental Evaluation}
\label{subsec:eval}
We assessed the representation performance of the proposed SimFLE after self-supervised pre-training. First, we assessed linear evaluation, semi-supervised and fine-tuning performance on AffectNet. We evaluated both in the case using 8 classes of AffectNet (AffectNet-8 or AN-8) and in the case using only 7 classes excluding the 'Contempt' class (AffectNet-7 or AN-7). Additionally, we also assessed performance on RAF-DB, FERPlus, and SFEW datasets to verify the transfer capability of SimFLE to other FER-W benchmarks. More details of each setup are in Supplement~\ref{app:details}. In all evaluation setup, we compare the performance of the proposed SimFLE and the other state-of-the-arts self-supervised methods. Additionally, the performance of fully supervised methods on ImageNet (Supervised-IN) or AffectNet (Supervised-AN) is presented as a baseline.

\paragraph{Linear Evaluation Protocol.} We evaluated the learned representations of the proposed SimFLE on AffectNet according to a linear evaluation protocol~\cite{SSLC3, SSLC5, SSLC8, SSLC9, SSLC10}. After pre-training, we froze the parameters of the backbone and train only the linear classifier added to the top of the backbone with the AffectNet labeled data (see Supplement~\ref{subapp:linear}). In Table.~\ref{table1}, the proposed SimFLE recorded 44.01\% and 50.40\% accuracy on AffectNet-8 and AffectNet-7, respectively. This is a noticeable improvement for both cases compared to other self-supervised methods and ImageNet-supervised baseline, indicating that our SimFLE method learns more decisive representations.

\paragraph{Semi-Supervised Learning.} We evaluated the proposed SimFLE method according to a semi-supervised setup~\cite{SSLC3, SSLC4, SSLC11, SSLC12} on AffectNet. After pre-training, we end-to-end trained the backbone with a linear classifier added to the top using only a subset of AffectNet labeled data (see Supplement~\ref{subapp:semi}). In Table.~\ref{table1}, the proposed SimFLE method recorded accuracies of 35.89\% and 46.19\% on AffectNet-8 when 1\% and 10\% of labeled data were used, respectively (AN-8$_{1}$ and AN-8$_{10}$). Additionally, SimFLE recorded accuracies of 40.23\% and 53.09\% on AffectNet-7 under the same conditions (AN-7$_{1}$ and AN-7$_{10}$). SimFLE outperformed the supervised baseline in all cases, indicating that the proposed SimFLE method is more effective than the supervised approach in the situation with limited labels. SimFLE also outperformed all other self-supervised methods.

\paragraph{Fine-tuning.} We evaluated the fine-tuning performance of SimFLE method on AffectNet. After pre-training, we end-to-end trained the backbone with a linear classifier added to the top on the full AffectNet labeled set (see Supplement~\ref{subapp:fine}). In Table.~\ref{table1}, SimFLE recorded accuracies of 52.61\% and 59.14\% on AffectNet-8 and AffectNet-7, respectively. SimFLE outperformed the supervised baseline, indicating that SimFLE has strengths in representation learning even without label data limitations. SimFLE also outperformed all other self-supervised methods.

\paragraph{Transfer Learning.} The purpose of self-supervised learning is to learn effective representations that can be transferred to other domains~\cite{SSLC5}. We evaluated the transfer performance of the proposed SimFLE on three other FER-W datasets: RAF-DB, FERPlus, SFEW. After pre-training, the backbone with a linear classifier added to the top was transferred on the three datasets, respectively (see Supplement~\ref{subapp:transfer}). In Table.~\ref{table1}, SimFLE recorded accuracies of 88.53\%, 88.11\%, and 56.38\%, respectively. The proposed SimFLE outperformed supervised baselines and other self-supervised methods, indicating that SimFLE learns more generalizable facial features.
\begin{table}[t]
\tablestyle{2pt}{1.1}
\begin{tabular}{l|x{30}x{30}x{30}x{30}x{30}}
\hline
\multicolumn{1}{c|}{\multirow{2}{*}{method}} & \multicolumn{5}{c}{accuracy (\%)}      \\ \cline{2-6} 
\multicolumn{1}{c|}{}                        & AN-8  & AN-7  & RAF-DB  & FER+  & SFEW  \\ \hline\hline
RAN~\cite{FS5}                               & 59.50$^\dagger$ & -               & 86.90 & 88.55 & 54.19 \\
SCN~\cite{FS11}                              & 60.23$^\dagger$ & -               & 88.14 & 88.01 & -     \\
KTN~\cite{FS15}                              & -               & 63.97$^\dagger$ & 88.07 & 90.49 & -     \\ 
EfficientFace~\cite{FS16}                    & 59.89$^\dagger$ & 63.70$^\dagger$ & 88.36 & -     & -     \\
LDL-ALSG~\cite{FS17}                         & -               & 59.35           & 85.53 & -     & 56.50 \\ 
MvT~\cite{FS18}                              & \textbf{61.40}$^\dagger$ & 64.57$^\dagger$ & 88.62 & 89.22 & -     \\
TransFER~\cite{FS12}                         & -               & 66.23$^\dagger$ & 90.91 & 90.83 & -     \\
APViT~\cite{FS19}                            & -               & \textbf{66.91}$^\dagger$ & \textbf{91.98} & \textbf{90.86} & \textbf{61.92} \\ \hline
SimFLE$_{\text{ResNet-50}}$                  & 59.21$^\dagger$ & 62.97$^\dagger$ & 88.53 & 88.11 & 56.38 \\
SimFLE$_{\text{ResNet-101}}$                 & \textbf{60.14}$^\dagger$     & \textbf{63.54}$^\dagger$     & \textbf{89.41} & \textbf{88.94} & \textbf{58.24}     \\ \hline
\end{tabular}
\caption{Comparison with state-of-the-art supervised methods. $^\dagger$Oversampling is used to alleviate class imbalance.}
\label{table2}
\end{table}

\paragraph{Comparison with State-of-the-arts.}
In Table.~\ref{table2}, we compared the proposed SimFLE with supervised state-of-the-art methods. Each performance of SimFLE is the result of the aforementioned fine-tuning or transfer setup. Oversampling was used for AffectNet-8 and AffectNet-7 according to standards in literature~\cite{FS5,FS16,FS19}. SimFLE shows competitive performance compared to supervised methods. This stands out when compared to methods that use smaller (ResNet-18)~\cite{FS5,FS11,FS15} or similar (ResNet-50, DeiT-S) backbones~\cite{FS16,FS17,FS18} . SimFLE lags behind methods that use huge backbones (ResNet-50+ViT-B)~\cite{FS12,FS19}, but the gap narrows when SimFLE uses a larger backbone (ResNet-101). It should be noted that SimFLE uses only a single network without additional modules for inference. Confusion matrices for each dataset are presented in Supplement~\ref{app:cf}.
\subsection{Ablation Study}
\label{sec:ablation}
We analyzed each component of the proposed SimFLE through the ablation study. The linear evaluation accuracy on AffectNet-8 was used for comparison.

\begin{figure}[t]
  \centering
   \includegraphics[width=0.48\linewidth]{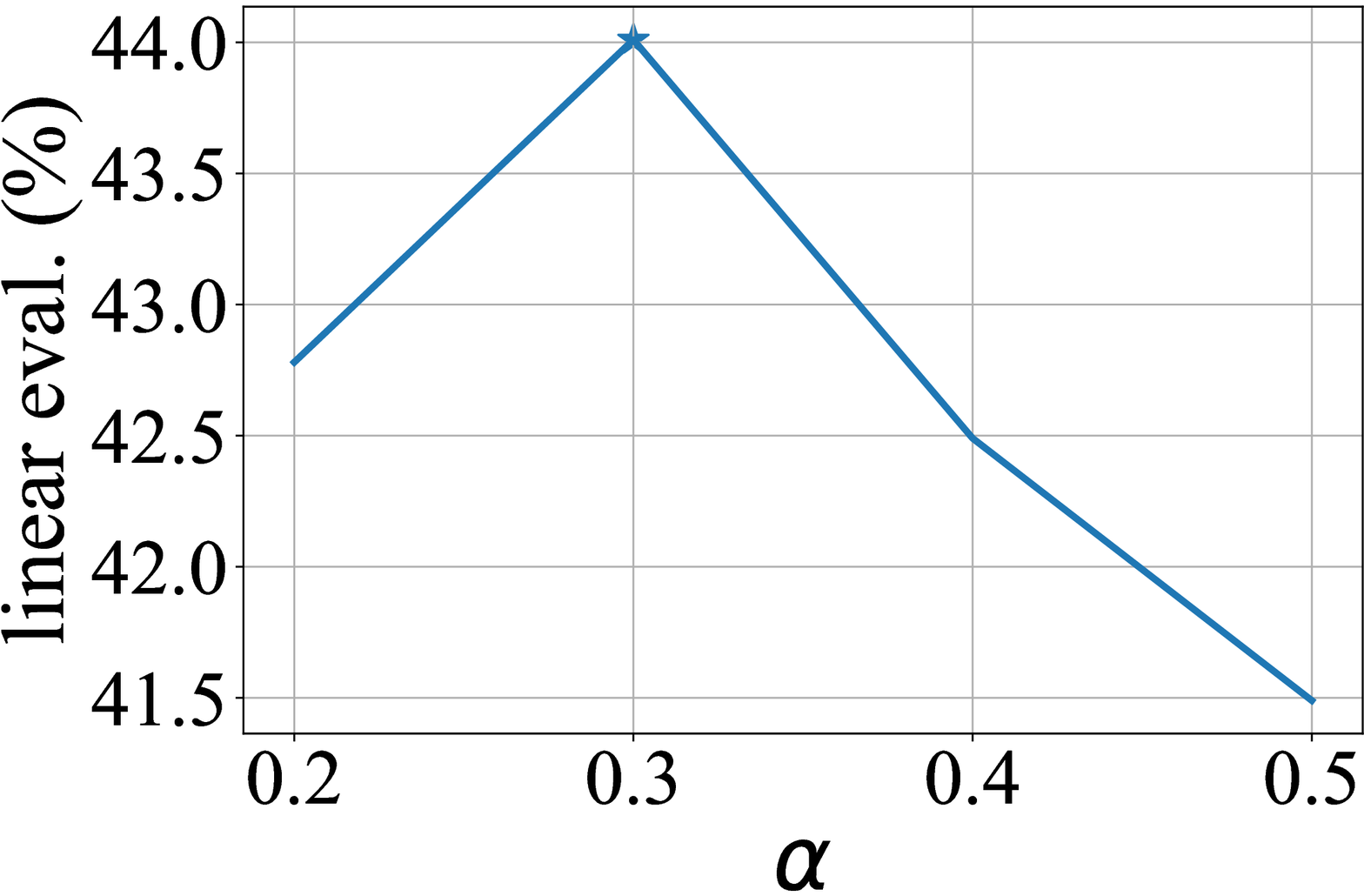}
   \includegraphics[width=0.48\linewidth]{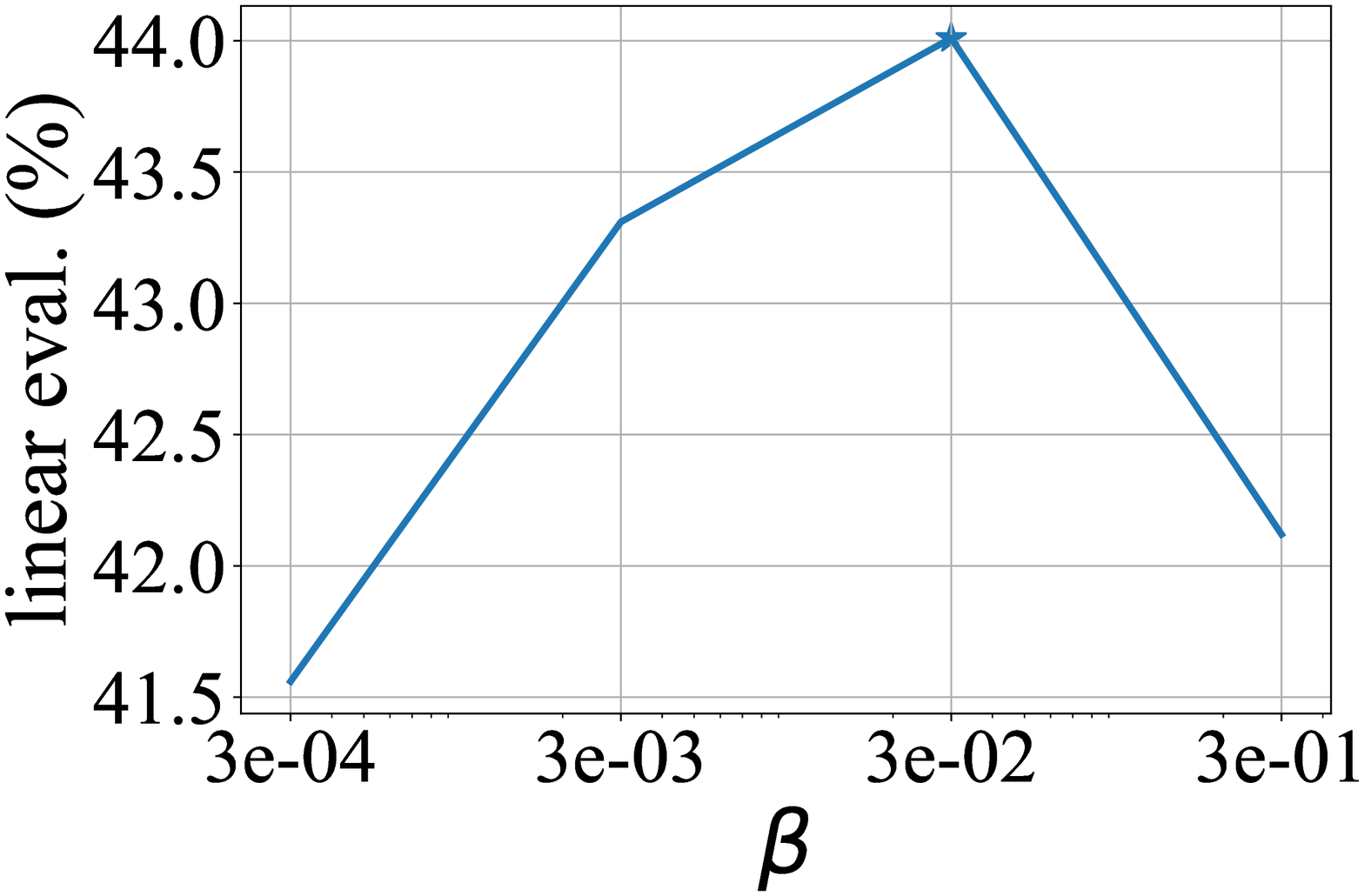}

   \caption{Ablation on $\alpha$ (left) and $\beta$ (right).}
   \label{fig5}
\end{figure}

\paragraph{Ablation on \boldmath$\alpha$\unboldmath~and \boldmath$\beta$\unboldmath}. First, we analyzed the effect of $\alpha$, the weight parameter of the $\mathcal{L}_{d}$. The results are shown in Fig.~\ref{fig5} left. $\alpha$ determines the strength of the connection. To explore the impact of $\alpha$, we evaluated the $\alpha$ from 0.2 to 0.5 with other parameters as default, and the optimal $\alpha$ value was 0.3 in our experiment. After determining the optimal value of $\alpha$, we analyzed the effect of $\beta$, the weight parameter of the $\mathcal{L}_{g}$. The results are shown in Fig.~\ref{fig5} right. $\beta$ determines the strictness of channel grouping. We evaluated the $\beta$ from 0.0003 to 0.3, and the optimal $\beta$ value was 0.03 in our experiment.
\begin{table}[t]
\tablestyle{4.0pt}{1.1}
\begin{tabular}{c|cccc|r|c}
\hline
index & GFL & FFL & masking & connection & FLOPs & linear eval. (\%) \\ \hline \hline
(a) & \checkmark & & & & 16.5 G & 41.64 \\
(b) & \checkmark & \checkmark & semantic & direct & 20.1 G & 20.98 \\
(c) & \checkmark & \checkmark & random & distil. & 20.1 G & 42.84 \\
(d) & \checkmark & \checkmark & semantic & distil. & 20.1 G & \textbf{44.01} \\ \hline
\end{tabular}
\caption{Ablation on each component of SimFLE.}
\label{table3}
\end{table}
\begin{figure*}[ht]
  \centering
   \includegraphics[width=1\linewidth]{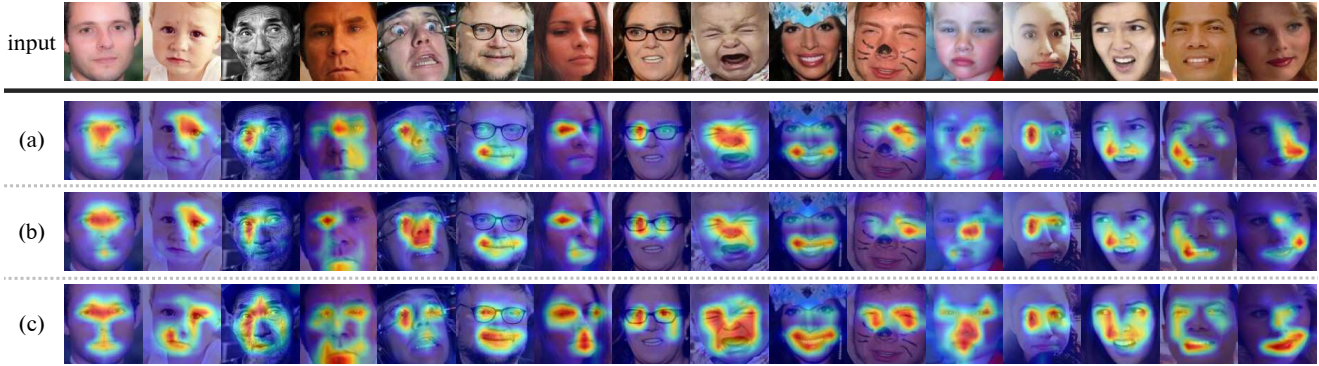}

   \caption{Visualization of attention maps of models trained with three different strategies. The input facial images are shown in the first row. (a) is attention maps when only GFL is used. (b) is attention maps when using GFL and FFL jointly, but using random masking for masked autoencoding. (c) is attention maps when using GFL and FFL jointly, but using semantic masking for masked autoencoding (our SimFLE). The visualization results show that our SimFLE has the strength to encode spatial and contextual properties of facial landmarks.}
   \label{fig6}
\end{figure*}

\paragraph{GFL and FFL.} We conducted an ablation study to analyze the effects of the components constituting SimFLE. The results are shown in Table.~\ref{table3}. First, we analyzed the effects of GFL and FFL. When only GFL was used (Table.~\ref{table3} (a)), the accuracy was $41.64\%$. This is the case where only global facial features are considered without using FFL for learning fine-grained facial landmarks. The result was lower than 44.01\% of the full architecture in which GFL and FFL are used jointly (Table.~\ref{table3} (d)), indicating that (1) global facial features alone are insufficient for satisfactory performance, and (2) the use of FFL to explore fine-grained facial landmarks results in performance improvements.

\paragraph{GFL Architecture.} To further prove the effect of FFL, we analyzed whether the addition of FFL had a consistent effect on different GFL architectures. Table.~\ref{table4} shows the linear evaluation accuracy when using different contrastive methods for SimFLE's GFL architecture. These changes are easy to implement because of SimFLE's modular design. In Table.~\ref{table4}, SimFLE consistently improves performance over each baseline, indicating that learning fine-grained facial landmarks is essential for better performance in FER-W.
\begin{table}[t]
\tablestyle{7.0pt}{1.1}
\begin{tabular}{lrcc}
\hline
\multicolumn{1}{c}{\multirow{2}{*}{method}} & \multicolumn{1}{c}{\multirow{2}{*}{FLOPs}} & \multicolumn{2}{c}{linear eval. (\%)} \\ \cline{3-4} 
\multicolumn{1}{c}{}                        & \multicolumn{1}{c}{}                       & AN-8       & AN-7       \\ \hline \hline
SimSiam~\cite{SSLC1}                        & 8.3 G                                     & 29.68             & 34.80             \\
SimFLE$_{\text{SimSiam}}$                   & 11.9 G                                    & \textbf{30.81}           & \textbf{35.66}              \\ \hline
MoCo~\cite{SSLC5}                           & 8.2 G                                     & 37.73             & 44.51             \\
SimFLE$_{\text{MoCo}}$                      & 11.8 G                                    & \textbf{39.96}             & \textbf{45.49}             \\ \hline
MoCov2~\cite{SSLC6}                         & 8.3 G                                     & 38.06             & 45.46             \\
SimFLE$_{\text{MoCov2}}$                    & 11.8 G                                    & \textbf{39.66}             & \textbf{47.43}             \\ \hline
SimCLR~\cite{SSLC4}                         & 4.1 G                                     & 38.41             & 44.97             \\
SimFLE$_{\text{SimCLR}}$                    & 7.7 G                                     & \textbf{40.54}             & \textbf{46.17}             \\ \hline
\end{tabular}
\caption{Linear evaluation accuracy when using different contrastive methods for SimFLE's GFL architecture.}
\label{table4}
\end{table}

\paragraph{Connection Strategy.} We analyzed the connection strategy of GFL and FFL. The accuracy was 20.98\% when the GFL and FFL were directly connected at the backbone level (Table.~\ref{table3} (b)). It was lower than 44.01\% when connected indirectly through knowledge distillation (Table.~\ref{table3} (d)). These results show that the direct connection between GFL and FFL makes the backbone training unstable. In FFL, FaceMAE refines detected features at the backbone level for easy reconstruction. On the other hand, in GFL, the backbone is adjusted to learn similarity according to the defined augmentation set. GFL and FFL use different learning objectives, and these conflicting interests can cause the backbone to confuse. Additionally, FaceMAE can cause corruption of the backbone through a semantic masking process. Therefore, we indirectly connected GFL and FFL through knowledge distillation so that we used both simultaneously while minimizing the risk of corruption.

\paragraph{Semantic Masking.} We analyzed the effect of the proposed semantic masking. When using random masking, the accuracy was 42.84\% (Table.~\ref{table3} (c)). Random masking also had some effect, but it was lower than using semantic masking (Table.~\ref{table3} (d)). In the case of random masking, a certain level of context can be considered because the masked facial image must be reconstructed with extremely limited information due to the high masking ratio~\cite{SSLG2, ETCM2}. However, it should be noted that the type and quality of the limited information provided to the encoder cannot be determined because the masked position is random. If limited information is given randomly, the model can learn encoding for unnecessary information such as background even if it learns some facial landmarks. On the other hand, the proposed semantic masking is based on semantic channel information and is trainable. Following an iterative training process, FaceMAE or backbone can refine the masking strategy to focus on the information most relevant to the facial landmarks. As shown in Fig.~\ref{fig4} (b), if the provided limited information is facial landmarks such as eyes and mouth, FaceMAE' encoder can convey to the backbone that how the spatial details of these facial landmarks should be encoded and how they are interrelated in the reconstruction process. For this reason, semantic masking has a key strength in that it can use the context learning ability of masked autoencoding and focus its effect on the fine-grained facial landmarks.
\subsection{Visualization}
\label{subsec:vis}
To further investigate the effectiveness of the proposed SimFLE, we explore the attention maps generated using Grad-CAM~\cite{ETCV1, ETCV2}. 

Fig.~\ref{fig1} shows the activation maps of SimFLE and other self-supervised methods (BYOL~\cite{SSLC3}, MoCov2~\cite{SSLC6}) for some examples. SimFLE excels at localizing facial landmarks when compared to other methods. Intriguingly, SimFLE localizes multiple facial landmarks simultaneously. This means that SimFLE can learn not only the spatial details of facial landmarks, but also the contextual relationships between them.

Fig.~\ref{fig6} shows the attention maps of the three models along with the input facial images. We pre-trained the backbone with three different training strategies, and each model was fine-tuned under the same conditions as described in Supplement~\ref{subapp:linear}. First, Fig.~\ref{fig6} (a) shows the attention maps when the backbone was pre-trained using only GFL (Table.~\ref{table3} (a)). The quality of facial landmarks localization is poor because the model does not have a strategy for fine-grained facial features. The localization quality is improved when designing an explicit learning strategy for facial landmarks by adding FFL as shown in Fig.~\ref{fig6} (b). However, the model in Fig.~\ref{fig6} (b) used random masking (Table.~\ref{table3} (c)), so the improvement is not significant. In some examples, the model fails to localize facial landmarks, or fails to capture associations between facial landmarks. On the other hand, the model in Fig.~\ref{fig6} (c) was trained with our SimFLE using a semantic masking strategy while jointly using GFL and FFL (Table.~\ref{table3} (d)). In this case, the model appears to focus better on facial landmarks than in the previous two cases. Intriguingly, it appears that the model understands the associations between the facial landmarks and uses it extensively in decisions, given that the model's attention region spans multiple facial landmarks compared to the previous two cases. These results show that FaceMAE using trainable semantic masking has strengths in learning the spatial details of facial landmarks and their associations. Extended visualization results are in the Supplement~\ref{app:vis}.
\section{Conclusion}
\label{sec:conclusion}
In this paper, we proposed SimFLE, a self-supervised method specialized for FER-W. Specifically, we proposed a novel FaceMAE module to learn the spatial details and contextual properties of fine-grained facial landmarks using elaborately designed trainable semantic masking. As well as learning global facial features, SimFLE learns decisive facial landmarks in an unsupervised way. Experimental results on several FER-W benchmarks show that our SimFLE outperforms the supervised baseline and other self-supervised methods, with strengths in localizing facial landmarks and identifying their associations.
%


{\small
\bibliographystyle{ieee_fullname}
\bibliography{main}
}

\clearpage
\newpage
\appendix

\section{Data Distribution}
\label{app:data}
\begin{table}[h]
\tablestyle{2.5pt}{1.1}
\begin{tabular}{l|cc|cc|cc|cc}
\hline
\multicolumn{1}{c|}{}      & \multicolumn{2}{c|}{AffectNet}       & \multicolumn{2}{c|}{RAF-DB} & \multicolumn{2}{c|}{FERPlus} & \multicolumn{2}{c}{SFEW} \\ \cline{2-9}
\multicolumn{1}{c|}{class} & train                        & test$^{\dagger}$  & train       & test        & train         & test         & train       & test$^{\dagger}$       \\ \hline
Neutral                    & 74,874                       & 500               & 2,524       & 680         & 8,733         & 1,083        & 144         & 84         \\
Happiness                  & 134,415                      & 500               & 4,772       & 1,185       & 7,284         & 892          & 184         & 72         \\
Sadness                    & 25,459                       & 500               & 1,982       & 478         & 3,022         & 382          & 161         & 73         \\
Surprise                   & 14,090                       & 500               & 1,290       & 329         & 3,136         & 394          & 94          & 56         \\
Fear                       & 6,378                        & 500               & 281         & 74          & 536           & 86           & 78          & 46         \\
Disgust                    & 3,803                        & 500               & 717         & 160         & 116           & 16           & 52          & 23         \\
Anger                      & 24,882                       & 500               & 705         & 162         & 2,098         & 269          & 178         & 77         \\
Contempt                   & 3,750                        & 500               & -           & -           & 120           & 15           & -           & -          \\ \hline
total                      &  287,651                     & 4,000             & 12,271      & 3,068       & 25,045        & 3,137        & 891         & 431        \\ \hline
\end{tabular}
\caption{Sample distribution of AffectNet (AffectNet-8), RAF-DB, FERPlus, and SFEW datasets. $^\dagger$ Validation sets are used for testing because test sets are not provided.}
\label{table5}
\end{table}
In Table.~\ref{table5}, we present the detailed sample distributions of AffectNet (AffectNet-8), RAF-DB, FERPlus, and SFEW. AffectNet-7 has a distribution excluding the `Contempt' class from AffectNet-8. Additionally, in the case of AffectNet and SFEW, validation sets are used for testing because test sets are not provided.

\section{Implementation Details}
\label{app:details}
\subsection{Linear Evaluation Protocol}
\label{subapp:linear}
After SimFLE pre-training, we froze the parameters of the backbone and added a linear classifier on top (batch statistics are also frozen). We only trained the classifier using the AffectNet labeled set. At training time, we applied random resized crops at $224\times224$ resolution and random horizontal flips as spatial augmentations. At test time, images were resized to $224\times224$ resolution and no other augmentation was used. In both cases, we normalize the color channels by subtracting the average color and dividing by the standard deviation (computed on AffectNet). We optimize the cross-entropy loss  for 100 epochs using SGD optimizer with momentum of 0.9 and weight decay of 0. The batch size was set to 1024, and the initial learning rate was set to 2.0. No learning rate decay schedule was used.
\subsection{Semi-Supervised Learning}
\label{subapp:semi}
After SimFLE pre-training, we added a linear classifier to the backbone and trained end-to-end using only a subset of AffectNet labeled data. Specifically, we used 1\% and 10\% of the AffectNet labeled set, respectively. At training time, we applied random resized crops at $224\times224$ resolution and random horizontal flips as spatial augmentations. At test time, images were resized to $224\times224$ resolution and no other augmentation was used. In both cases, we normalize the color channels by subtracting the average color and dividing by the standard deviation (computed on AffectNet). We optimize the cross-entropy loss  for 100 epochs using SGD optimizer with momentum of 0.9 and weight decay of  0.0001. The batch size was set to 512, and the initial learning rate was set to 0.05. The learning rate had a cosine decay schedule~\cite{ETCE2, SSLC4}.
\subsection{Fine-tuning}
\label{subapp:fine}
After SimFLE pre-training, we froze the parameters of the backbone and added a linear classifier on top (batch statistics are also frozen). First, only the linear classifier was trained as in \ref{subapp:linear}. After that, the network was trained end-to-end on the full AffectNet labeled set. At training time, we applied random resized crops at $224\times224$ resolution and random horizontal flips as spatial augmentations. At test time, images were resized to $224\times224$ resolution and no other augmentation was used. In both cases, we normalize the color channels by subtracting the average color and dividing by the standard deviation (computed on AffectNet). We optimize the cross-entropy loss  for 100 epochs using SGD optimizer with momentum of 0.9 and weight decay of  0.0001. The batch size was set to 512, and the initial learning rate was set to 0.001. The learning rate had a cosine decay schedule~\cite{ETCE2, SSLC4}.
\subsection{Transfer Learning}
\label{subapp:transfer}
After SimFLE pre-training, we froze the parameters of the backbone and added a linear classifier on top (batch statistics are also frozen). First, only the linear classifier is trained as in \ref{subapp:linear}.  Since RAF-DB and SFEW consist of only 7 emotion classes, the linear classifier was trained on AffectNet-7 in this case. After that, the network was trained end-to-end on three other FER-W datasets: RAF-DB, FERPlus, SFEW. At training time, we applied random resized crops at $224\times224$ resolution and random horizontal flips as spatial augmentations. At test time, images were resized to $224\times224$ resolution and no other augmentation was used. In both cases, we normalize the color channels by subtracting the average color and dividing by the standard deviation. For RAF-DB and SFEW, values computed from AffectNet were used, and for FERPlus, values recalculated from FERPlus were used. We optimize the cross-entropy loss for 100 epochs using SGD optimizer with momentum of 0.9 and weight decay of  0.0001. The batch size was set to 512, and the initial learning rate was set to 0.001. The learning rate had a cosine decay schedule~\cite{ETCE2, SSLC4}.
\section{Confusion Matrix}
\label{app:cf}
\begin{figure*}[h]
  \centering
   \includegraphics[width=0.8\linewidth]{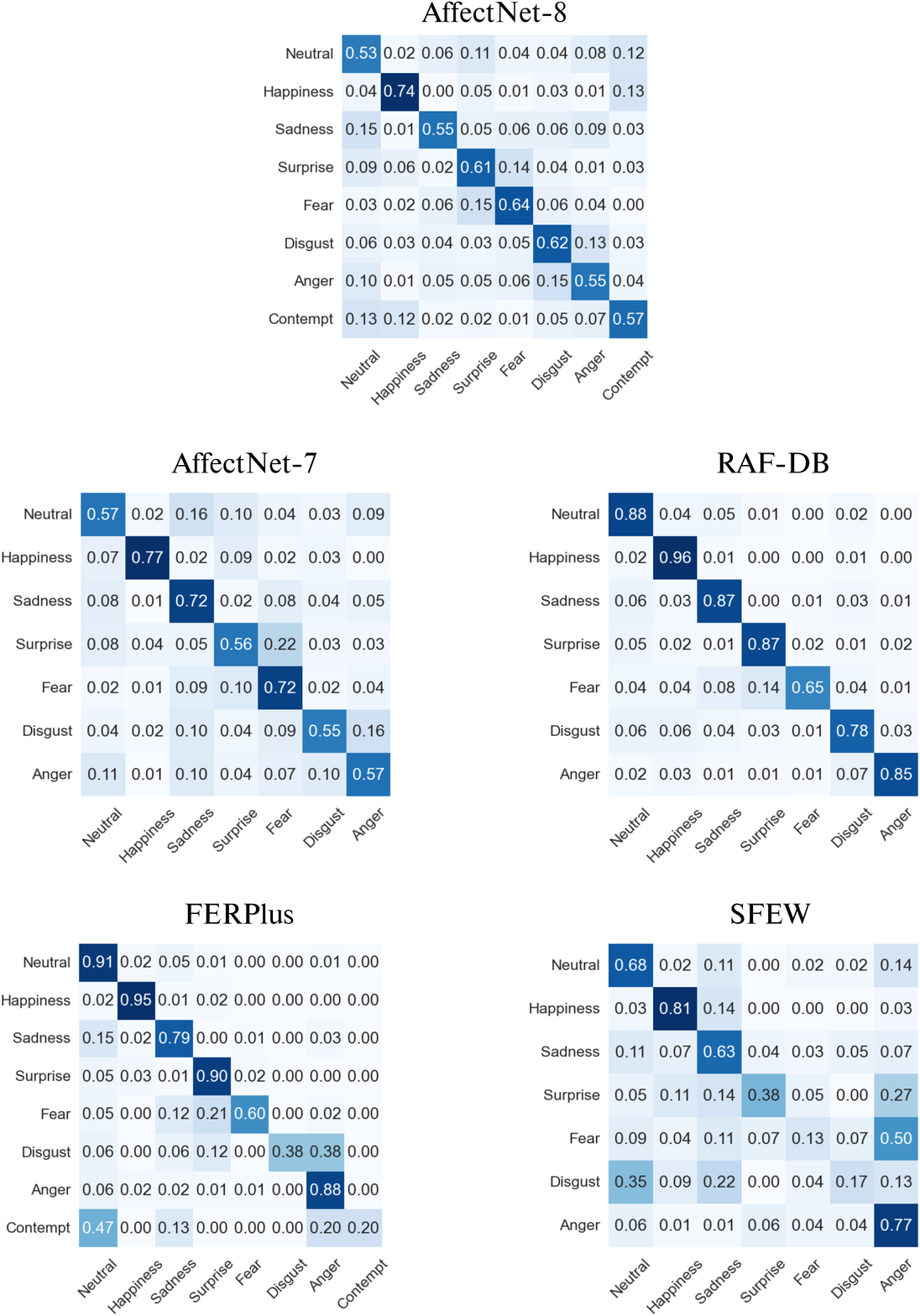}

   \caption{Confusion matrices for AffectNet-8, AffectNet-7, RAF-DB, FERPlus and SFEW. These were calculated based on the SimFLE with ResNet-101 backbone in Table.~\ref{table2}.}
   \label{fig7}
\end{figure*}

In Fig.~\ref{fig7}, we present the confusion matrices of SimFLE with ResNet-101 backbone in Table.~\ref{table2}. The results are presented for AffectNet-8, AffectNet-7, RAF-DB, FERPlus and SFEW. For all datasets, the `Happiness' class has the highest accuracy. Additionally, `Happiness-Contempt', `Surprise-Fear', and `Disgust-Anger' are mainly confused.
\section{Visualization}
\label{app:vis}
\begin{figure*}[h]
  \centering
   \includegraphics[width=1\linewidth]{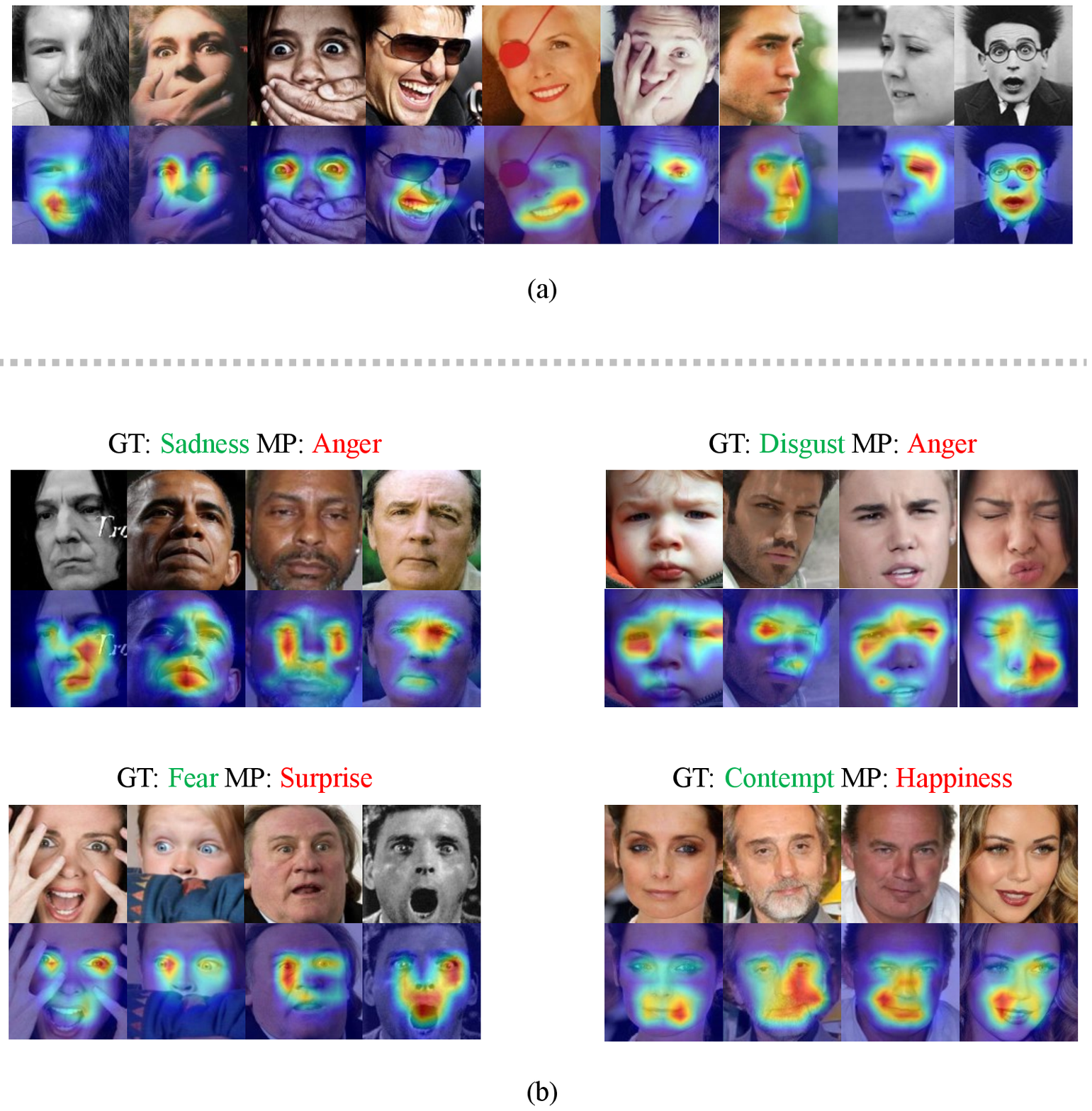}

   \caption{Visualization of attention maps of (a) more challenging examples and (b) misclassified cases. (a) SimFLE-trained model performed well for more challenging examples, such as rotated faces or occlusion of some of the facial landmarks. (b) Apart from the facial landmarks localization results, ground-truth lables (GT) and model predictions (MP) are mainly misclassified if they have similar visual semantics.}
   \label{fig8}
\end{figure*}

In Fig.~\ref{fig8}, we present extended visualization results of (a) more challenging examples and (b) misclassified cases.

Fig.~\ref{fig8} (a) shows attention maps when the SimFLE-trained model is fed more challenging examples where some of the facial landmarks are occluded, the face is rotated, or not tightly captured. Even if some of the facial landmarks are occluded, the SimFLE-trained model identifies visible facial landmarks and optionally localizes those regions (the first to sixth columns in Fig.~\ref{fig8} (a)). This is similar when the face is rotated to limit visible facial landmarks (the seventh to eighth columns in Fig.~\ref{fig8} (a)). The results for the rotated face also show that the SimFLE-trained model is somewhat invariant to the deformation of facial landmarks. Finally, the SimFLE-trained model is able to localize facial landmarks even when the size of the facial landmarks is relatively small because the face is not tightly captured in the input image (the ninth column of Fig.~\ref{fig8} (a)). 

Fig.~\ref{fig8} (b) shows the attention maps for the four cases in which the model mainly misclassifies. Ground-truth labels (GT) and model prediction labels (MP) are presented together. The model mainly misclassifies emotion classes that have similar visual semantics. For example, `Fear' and `Surprise' are often accompanied by a screaming mouth and widened eyes. Similarly, `Happiness' and `Contempt' can have similar visual semantics of smile and sneer. This can confuse classification, regardless of how well the model localizes facial landmarks.
%

\newpage

%
\end{document}